\documentclass[letterpaper, 10 pt, conference]{ieeeconf}  %
\IEEEoverridecommandlockouts                              
\usepackage{graphicx} 
\usepackage{amsmath} 
\usepackage[]{algorithm2e}
\usepackage{amssymb}  
\usepackage{booktabs}
\usepackage{cite}
\usepackage{lipsum}
\usepackage{xcolor}

\newcommand{\vect}[1]{\mathbf{ #1}}

\newcommand{\vb}{\vect{b}}

\newcommand{\vs}{\vect{s}}

\newcommand{\cO}{\mathcal{O}}

\usepackage{flushend}

\title{\LARGE \bf Evaluating Merging Strategies for Sampling-based Uncertainty Techniques in Object Detection}
\author{Dimity Miller, Feras Dayoub, Michael Milford, and Niko S{\"u}nderhauf%
\thanks{The authors are with the ARC Centre of Excellence for Robotic Vision, Queensland University of Technology (QUT), Brisbane, Australia. Contact: dimity.miller@hdr.qut.edu.au}
\thanks{This research was conducted by the Australian Research Council Centre of Excellence for Robotic Vision (project number CE140100016). Michael Milford is supported by an Australian Research Council Future Fellowship (FT140101229).}
}
\date{August 2018}

\begin{document}

\maketitle

\begin{abstract}
There has been a recent emergence of sampling-based techniques for estimating epistemic uncertainty in deep neural networks. While these methods can be applied to classification or semantic segmentation tasks by simply averaging samples, this is not the case for object detection, where detection sample bounding boxes must be accurately associated and merged. A weak merging strategy can significantly degrade the performance of the detector and yield an unreliable uncertainty measure. This paper provides the first in-depth investigation of the effect of different association and merging strategies. We compare different combinations of three spatial and two semantic affinity measures with four clustering methods for MC Dropout with a Single Shot Multi-Box Detector. Our results show that the correct choice of affinity-clustering combination can greatly improve the effectiveness of the classification and spatial uncertainty estimation and the resulting object detection performance.
We base our evaluation on a new mix of datasets that emulate near open-set conditions (semantically similar unknown classes), distant open-set conditions (semantically dissimilar unknown classes) and the common closed-set conditions (only known classes).

\end{abstract}

\section{Introduction}
Convolutional Neural Networks (CNNs) are widely recognized as unable to measure their lack of knowledge, or \emph{uncertainty}\cite{nguyen2015deep}. While this does not pose an issue in closed-set conditions, where testing is \emph{in distribution} with the training environment, a serious decrease in performance is observed in \emph{open-set conditions}~\cite{torralba2011unbiased, hendrycks2017baseline}, where object classes not present during training are encountered and testing is \emph{out of distribution} with the training environment ~\cite{scheirer2013toward}. With the current push towards widespread and accessible autonomous systems that must operate in the open-set real world, there is a strong motivation to ensure that deep learned systems are robust and capable of expressing uncertainty. 

\begin{figure}[t]
    \centering
    \includegraphics[width=1\linewidth,clip=true]{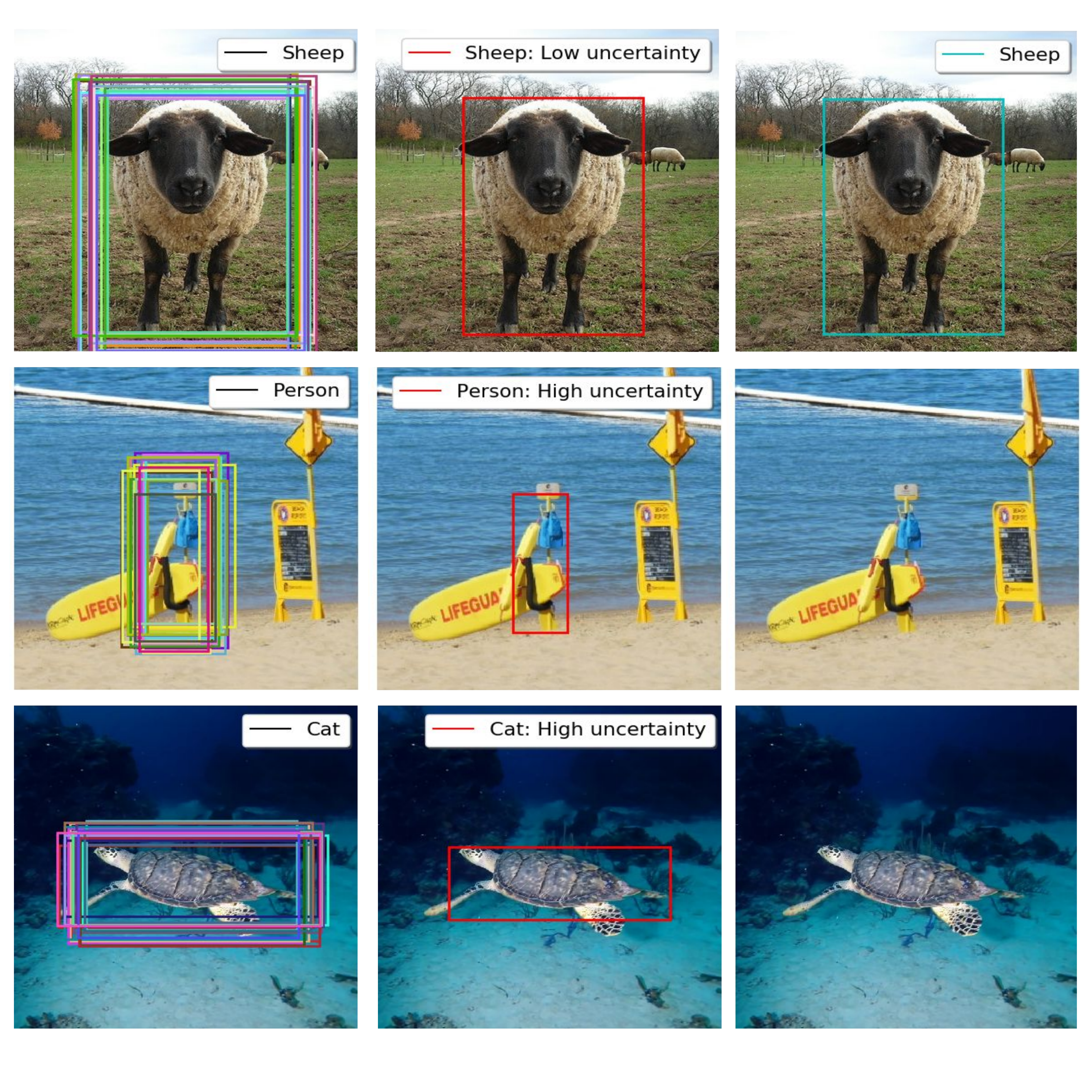}
    \vspace{-20pt}
    \caption{With a high performing association method, an object detection network using a sampling-based uncertainty technique can cluster raw samples (left column) into observations of objects with a reliable uncertainty measurement (middle column) and use this uncertainty to accept correct detections and reject incorrect detections (right column). The detector must be able to do this reliably in a closed-set environment (top row) and in a variety of open-set environments (middle, bottom rows). Results in this figure demonstrate the performance of our best performing merging strategy.}
    \vspace{-14pt}
    \label{fig:firstpage}
\end{figure}

Recent techniques used to quantify uncertainty in CNNs are often sampling-based, where stochasticity is introduced into the network, multiple samples are extracted, and an uncertainty measurement is approximated from the variation between samples \cite{gal2016dropout, lakshminarayanan2017simple, azizpour2018bayesian} (see Fig. \ref{fig:firstpage}). While for classification and segmentation tasks samples can simply be averaged, this is not a trivial task for object detection, where diverse detection samples must be correctly associated and merged for multiple objects in an image. 

While previous works have integrated sampling-based techniques into object detection networks \cite{miller2018dropout, liang2017enhancing}, they did not address how their proposed association and merging strategy affected the performance of the object detector and the effectiveness of the measured uncertainty.
The contributions of this paper are as follows:
\begin{itemize}
\item We provide the first in-depth investigation of the effect of affinity measures and clustering techniques for associating bounding box samples on uncertainty effectiveness and object detection performance (using MC Dropout with a Single Shot  Multi-Box  Detector, SSD).
\item We establish an evaluation protocol and metrics targeting the quality of both spatial and classification uncertainty for object detection and highlight the metrics that are most relevant for robotic applications.
\item We propose an association and clustering strategy that provides the most effective uncertainty measure, particularly for robotic applications.
\end{itemize}
\begin{figure*}[t]
    \centering
    \includegraphics[width=1\linewidth]{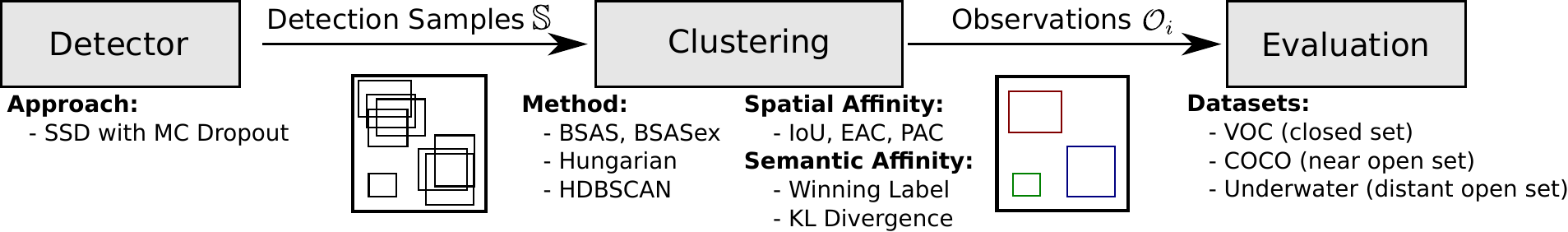}
    \caption{Sampling-based approaches such as SSD with Monte Carlo Dropout~\cite{miller2018dropout} are state of the art for obtaining uncertainty estimates for visual object detection. Such approaches produce large numbers of detections that need to be clustered into \emph{observations} of individual objects. We analyse the performance of four different clustering methods in combination with three spatial and two semantic affinity measures. Using three different datasets that represent closed set, near open, and distant open set conditions, allows us to evaluate how well different detector-clustering-affinity combinations avoid erroneous detections of unknown objects, while maintaining good detection performance on known objects. Good performance under such challenging conditions is essential for many robotics application that require object detection in uncontrolled real-world environments. }
    \label{fig:overview}
\end{figure*}

The remainder of the paper is structured as follows: Section \ref{sec:lit} outlines the related work on sampling-based uncertainty and Section \ref{sec:od} describes sampling-based uncertainty techniques in object detectors. Section \ref{sec:methods} outlines the affinity measures and clustering techniques that were tested for associating detections. Section \ref{sec:evaluation} describes the experimental evaluation and metrics and Section \ref{sec:results} presents the experimental results. Finally,  Section \ref{sec::conclusions} draws conclusions and suggests areas for future work.

\section{Related Work}
\label{sec:lit}
Visual object detection is the process of localizing all instances of known object classes in an image with tight bounding boxes, and assigning the correct class label to them. For a very comprehensive survey on deep learning based methods that dominate the state of the art, we refer the reader to \cite{liu2018survey}. In contrast to earlier methods such as R-CNN~\cite{R-CNN}, FasterR-CNN~\cite{FasterR-CNN} integrated the process of region proposal generation as a branch in the network itself. More current methods such as Single  Shot  Multi-Box  Detector (SSD)~\cite{SSD} or YOLO~\cite{YOLO,Yolo9000} took the idea further and unified the detection and proposal generation into one branch in the network. The object detection approach used in this paper is SSD~\cite{SSD}.

A common assumption in deep learning is that trained models will be deployed under \emph{closed-set} conditions \cite{bendale2015towards,torralba2011unbiased}, i.e. the classes encountered during deployment are known in advance, and are the same as during training. However, robots often operate in open-set conditions \cite{scheirer2013toward, bendale2015towards}, where they inevitably encounter objects that were not part of the training data.

Current attempts at improving open-set performance of deep learning systems have focused on the estimated uncertainty in the network predictions, utilizing
calibration techniques \cite{hendrycks2017baseline,guo2017calibration}, Bayesian deep learning \cite{mackay1992practical, neal1995bayesian} with approximations such as Monte Carlo (MC) Dropout~\cite{gal2016dropout, kendall2017uncertainties} and MC Batch Normalization ~\cite{azizpour2018bayesian}, or Deep Ensemble methods \cite{lakshminarayanan2017simple}. 

MC Dropout has previously been used to estimate the epistemic uncertainty for regression, image classification and segmentation ~\cite{kendall2017uncertainties, kendall2016bayesian,kendall2016modelling}. Applying MC Dropout or other sampling-based techniques (such as ensembles or MC Batch Normalization) to object detection is non-trivial as it requires the partitioning of individual detections using some kind of affinity-based clustering method. This has only recently been demonstrated for the first time for object detection under open-set conditions in~\cite{miller2018dropout}, and for vehicle detection in 3D Lidar data by~\cite{feng2018lidar}. Neither \cite{miller2018dropout, feng2018lidar} nor other work in progress~\cite{gurau2018dropout} analysed the influence of different affinity measures and clustering methods on the object detection performance under open-set conditions. Our paper is the first to deliver this important evaluation.

\section{Sampling-based Object Detection}
\label{sec:od}
A single forward pass through an object detector produces a set of individual detections $S = \{D_1, \dots, D_k\}$. Each detection  $D_i=\{\vb, \vs \}$ comprises bounding box coordinates  $\vb = (x_1, y_1, x_2, y_2)$ and a distribution of softmax scores for the $m$ known classes $\vs = (s_1, \dots, s_m)$. The winning class and score for each detection correspond to the maximum softmax score in $\vs$. 

Sampling-based methods~\cite{lakshminarayanan2017simple, miller2018dropout, azizpour2018bayesian} combined with object detectors produce a set of $n$ samples $\mathbb{S} = \{S_1, \dots, S_n\}$ where each sample $S_j = \{D_1, \dots, D_k\}$ in turn contains a set of detections $D_i$ as defined above.

Given the set of samples $\mathbb{S}$, all detections $D_i$ must be grouped to form \emph{observations} $\cO_i$ of objects in the scene~\cite{miller2018dropout}. This process is illustrated in Fig.~\ref{fig:overview} and can be decomposed into three steps:
\begin{enumerate}
    \item measuring the affinity (similarity) between detections
    \item clustering detections into groups based on their affinity
    \item forming an observation from the clustered detections
\end{enumerate}

For the last step we follow the approach detailed in \cite{miller2018dropout}, where groups of 2 or more detections are converted to observations by averaging the bounding boxes $\vb$ and softmax score distributions $\vs$.

For the remainder of the paper we focus on analysing the influence of different affinity measures (step 1) and clustering techniques (step 2) on the overall performance of the object detector. Such an analysis has not yet been performed in the literature, but is an important step towards developing robust visual object detectors that are reliable in open-set conditions for robotic applications.

\section{Evaluated Affinity and Clustering Methods}\label{sec:methods}
We analyse and compare the performance of different combinations of spatial affinity measures, 
semantic affinity measures 
and clustering methods. 
This section describes the evaluated approaches, before detailing the experimental setup in section \ref{sec:experiments} and presenting results in Section \ref{sec:results}.

\subsection{Affinity Measures}
\textbf{Spatial Affinity:} To measure the spatial affinity between detections, the location and shape of bounding boxes can be compared. Intersection Over Union (IoU) is a well established method for measuring spatial affinity, and a Product Association Cost (PAC) \cite{sanchez2016online} and Exponential Association Cost (EAC) (referred to as motion affinity and shape affinity in previous work \cite{yu2016poi}) have also been demonstrated in object tracking literature.

\textbf{Semantic affinity:} A broad measure of the semantic affinity between detections is whether they have the Same winning class Label (SL). A comprehensive measure of semantic affinity is the KL divergence (KL), which measures the difference between the distributions of softmax scores of detections.
\subsection{Clustering Techniques}
The clustering techniques used can be distinguished by their \emph{intra-sample exclusivity}. We define intra-sample exclusivity as excluding detections from the same sample $S_i$ from being allocated to the same observation $\cO_i$. 
 
 \subsubsection{Basic Sequential Algorithmic Scheme (BSAS)} BSAS~\cite{theodoridis2003konstantinos} is a basic clustering algorithm that sequentially groups detections that meet a minimum threshold for affinity, $\theta$. For each detection, if the maximum affinity between the detection $D_i$ and any existing cluster $C_j$ meets $\theta$, the detection joins the cluster - otherwise the detection is allocated to a new cluster. This method has previously been applied successfully to MC Dropout with a SSD detector \cite{miller2018dropout}, though only one affinity threshold $\theta$ was tested.
 
IoU can be used as a spatial affinity measure, with various minimum IoU thresholds, $\theta$. This can be combined with SL affinity (IoU \& SL) by only comparing detections and clusters with the same winning label. An IoU \& KL variation can be tested by adjusting the affinity measure to $IoU - KL$ and the affinity threshold to IoU $ \theta - 0.1$. 

\subsubsection{BSAS with intra-sample exclusivity (BSAS excl.)}
While BSAS allows for intra-sample clustering, it can be adapted to be intra-sample exclusive. This is implemented by only comparing detections to clusters that do not contain detections from the same sample. For this clustering method, the IoU \& KL affinity was altered to group a detection with the minimum $KL$ cost cluster that also meets the IoU $\theta$. 

 \subsubsection{Hungarian Method}
The Hungarian Method is an established optimization algorithm \cite{kuhn1955hungarian} that solves an  $m \times n$ assignment problem. To use this algorithm for intra-sample exclusive clustering, we form an initial set of clusters from the detections from the first sample $\mathbb{S}_1$. We then sequentially compute the cost matrix between $m$ detections from a $\mathbb{S}_i$ and $n$ existing clusters. If there are more detections than clusters, $m > n$, new clusters are created from the unassigned detections. 

The spatial cost can be calculated as the negation of IoU, PAC or EAC between detections and a cluster. SL affinity can be incorporated by assigning near infinite cost between detections and clusters with different winning class labels. A KL variation can also be tested by adding a KL cost matrix to the existing spatial cost matrix.

 \subsubsection{Hierarchical Density-Based Spatial Clustering of Applications with Noise (HDBSCAN)} HDBSCAN is an extension of DBSCAN, a density-based clustering algorithm, that is robust to parameter selection and allows for clusters of varying densities \cite{McInnes2017}. Unlike other density-based clustering methods, such as K-Means, it does not require an estimated number of clusters and is robust to noisy data, thus making it suitable for this application. HDBSCAN calculates its own affinity measures when given two-dimensional data as input. Given the two-dimensional limit on input data, only spatial data was provided to the algorithm, including bounding box centroids (Centroid), top-left corner coordinates (Corner) and euclidean distance between corners and the image boundaries (Euclidean).

\section{Evaluation}
\label{sec:evaluation}
\subsection{Experimental Setup}
\label{sec:experiments}
\textbf{Using Uncertainty in Object Detection}: Given our observations $\cO_i$ from the clustering methods outlined in Section \ref{sec:methods}, we can extract uncertainty and use this to improve performance in open-set conditions.

Classification uncertainty is extracted from an observation by taking the entropy of the final softmax score distribution. Spatial uncertainty is extracted from an observation by measuring the total variance of bounding box coordinates within an observation's cluster of detections (as proposed by \cite{feng2018lidar}).

If $U(\cO_i)$ represents the uncertainty extracted from an observation, and $\delta$ is a given uncertainty threshold, we can then \textbf{reject detections with a high uncertainty} $U(\cO_i) > \delta$, and \textbf{accept detections with a low uncertainty} $U(\cO_i) \leq \delta$. For best object detection performance, a detector should accept all correct detections and reject all incorrect detections.  A good uncertainty technique and clustering method will assign a high uncertainty to incorrect detections and a low uncertainty to correct detections. An incorrect detection can include inaccurate detections of a known class (we call these closed-set errors) or a detection of an unknown class (we call these open-set errors).

\textbf{Clustering Techniques and Affinity Measures:} As a baseline, the standard object detector (SSD) with no sampling is tested. The clustering method used in \cite{miller2018dropout} (BSAS IoU 0.95) is also used as a baseline. These baselines are compared to the following clustering variations:
\begin{enumerate}
    \item \textbf{BSAS}: thresholds $\theta$ = \{0.7, 0.8, 0.9, 0.95\}, affinity = \{IoU, IoU \& SL and IoU \& KL\}
    \item \textbf{BSAS excl.}: thresholds $\theta$ = \{0.7, 0.8, 0.9, 0.95\}, affinity = \{IoU, IoU \& SL and IoU \& KL\}
    \item \textbf{Hungarian}: affinity = \{IoU, IoU \& SL, IoU \& KL, Product, Product \& SL, Product \& KL, Exponential, Exponential \& SL, Exponential \& KL\}
    \item \textbf{HDBSCAN}: input data = \{Centroid, Corner, Euclidean\}
\end{enumerate}

\textbf{Object Detectors and Uncertainty Techniques:}
For the following experiments, we used MC Dropout as our uncertainty technique with SSD as our object detector.

We implement MC Dropout into SSD with the following method, which was adapted from \cite{miller2018dropout}: 1) SSD is trained on PASCAL VOC 2007 and 2012 with a VGG16 base network, 2) this SSD is then fine-tuned with dropout layers on the final two convolutional layers of the VGG16 component of SSD and a dropout probability of 0.5. 20 samples were obtained by completing 20 forward passes through MC Dropout SSD with the dropout layers enabled. We only accept detections where the class label corresponds to the winning class from the softmax score distribution.

\subsection{Performance Metrics}
\label{sec:metrics}
We now define metrics to measure both \emph{object detection performance} and \emph{uncertainty effectiveness}. Object detection performance represents how well a detector correctly classifies and localises to known objects in a scene, while uncertainty effectiveness demonstrates how accurately an uncertainty measure can be used to distinguish between correct and incorrect detections. The two key metrics to represent the overall performance are Mean Average Precision (mAP) and Uncertainty Error (UE). Three auxilliary metrics (Area under ROC and PR curves, and IoU with the ground truth) allow 
an in-depth evaluation of uncertainty effectiveness \cite{hendrycks2017baseline, devries2018learning, liang2017enhancing}.
\subsubsection{\textbf{Mean Average Precision (mAP)}} is an accepted metric for evaluating object detection performance \cite{everingham2010pascal, lin2014microsoft}. mAP measures a detector's ability to detect all objects in a closed-set dataset with a correct classification and accurate localization (IoU $\geq$ 0.5), while minimizing incorrect detections and their confidence score. A perfect mAP score is 100\%.

\subsubsection{\textbf{Uncertainty Error (UE)}} also called detection error, has previously been used in literature to evaluate uncertainty effectiveness for classification tasks \cite{devries2018learning, liang2017enhancing}. This metric represents the ability of an uncertainty measure to accept correct detections and reject incorrect detections (where an incorrect detection can include closed-set errors and/or open-set errors). The uncertainty error, defined in Eq.(\ref{eq:ue}), is the probability that a detection is incorrectly accepted or rejected at a given uncertainty threshold ($\delta$), i.e. the proportion of correct detections $\mathbb{D}_c$ that are incorrectly rejected and the proportion of incorrect detections $\mathbb{D}_i$ that are incorrectly accepted. 
\begin{equation}
   UE(\delta) = 0.5 \frac{|U(\mathbb{D}_c) > \delta |}{|\mathbb{D}_c|} + 0.5 \frac{|U(\mathbb{D}_i) \leq \delta |}{|\mathbb{D}_i|}
    \label{eq:ue}
\end{equation}

The minimum uncertainty error represents the ideal threshold for separating correct and incorrect detections and the highest uncertainty effectiveness achievable by a detector. Perfect performance is an uncertainty error of 0\%, where all $\mathbb{D}_c$ are accepted, and all $\mathbb{D}_i$ are rejected. This metric weights the probability of correct detections and incorrect detections equally.
 
\subsubsection{Area Under the ROC curve (AUROC)} represents the probability that a correct detection has a lower uncertainty than an incorrect detection \cite{fawcett2006introduction}, where a perfect AUROC score is 100\%. It is calculated by finding the area under the ROC curve, where we define a true positive as a correct detection $D_c$ that is accepted, and a false positive as an incorrect detection $D_i$ that is accepted. 

\subsubsection{Area Under the Precision Recall curve (AUPR)} measures uncertainty effectiveness similarly to AUROC, though it can be more reliable when the number of positive examples and negative examples are highly contrasting \cite{saito2015precision}. Calculated by finding the area under the precision-recall curve, this metric is influenced by the defined positive class. We define AUPR-In, where the positive class is accepted detections ($D \leq \delta$) and AUPR-Out, where the positive class is rejected detections ($D > \delta$). AUPR-In represents the ability of a detector to accept all correct detections, while minimizing the number of accepted incorrect detections. AUPR-Out represents the ability of a detector to reject all incorrect detections, while minimizing the number of rejected correct detections. For a robotics context, where systems often act upon the real world based on their perception, it is often more important to remove all incorrect detections, and therefore AUPR-Out is the more informative metric. In each case, a perfect AUPR score is 100\%.

\subsubsection{IoU with ground truth objects (GT-IoU)} measures the spatial accuracy of an object detector. It is calculated by finding the maximum IoU between a detection and a GT object of the same class. Best performance is a GT IoU of 1 for all detections.

\subsection{Datasets}
\label{sec:datasets}
Each affinity-clustering combination was evaluated on a range of datasets that contained both closed-set and open-set conditions: 
\subsubsection{Closed-Set Data} contains 4952 images  from the PASCAL VOC 2007 test dataset \cite{everingham2010pascal} and was used to evaluate the closed-set performance of the detector, as PASCAL VOC data was used for training.
\subsubsection{Near Open-Set Data} contains 918 images from the COCO 2017 validation dataset \cite{lin2014microsoft} that do not contain any instances of the 20 PASCAL VOC classes known to the detectors. This data simulates \emph{near} open-set conditions, as while only unknown classes are present, the visual appearance and semantic information is very similar to the training data.
\subsubsection{Distant Open-Set Data} consists of 2899 images taken from varying underwater scenes. The images contain various marine life and underwater artefacts, and no instances of the detector known classes. Visually, and contextually, this data is very different to the training data, and thus we refer to it as \emph{distant} open-set conditions.

\section{Results}
\label{sec:results}

\begin{table*}[tb]
\centering
\caption{The uncertainty effectiveness for merging techniques can be decomposed into performance on closed-set data, distant open-set data and near open-set data. Arrows indicate direction of better performance.}
\label{tab:performance_detail}
\scalebox{0.85}{
\setlength\tabcolsep{2.2pt}
\begin{tabular}{@{}lcccc|cccc|cccc|cccc@{}}
 \toprule
 
 \textbf{Datasets:}& \multicolumn{4}{c}{\textbf{Closed-Set}} & \multicolumn{4}{c}{\textbf{Closed-Set \& Distant Open-Set}} & \multicolumn{4}{c}{\textbf{Closed-Set \& Near Open-Set}}  & \multicolumn{4}{c}{\textbf{All}} \\ 
  \multicolumn{5}{r}{(Correct Detections \& Closed-Set Error)} & \multicolumn{4}{c}{(Correct Detections \& Distant OSE)} & \multicolumn{4}{c}{(Correct Detections \& Near OSE)}  & \multicolumn{4}{c}{(All detections)} \\ \midrule
 & UE (maP) &  AUROC & AUPR& AUPR & UE (maP) &  AUROC & AUPR& AUPR & UE (maP) &  AUROC & AUPR & AUPR & UE (maP) & AUROC & AUPR & AUPR  \\
   &	$\downarrow$ ($\uparrow$) & $\uparrow$ & In $\uparrow$ & Out 	$\uparrow$   &	$\downarrow$ ($\uparrow$) & $\uparrow$ & In $\uparrow$ & Out 	$\uparrow$   &	$\downarrow$ ($\uparrow$) & $\uparrow$ & In $\uparrow$ & Out 	$\uparrow$  &	$\downarrow$ ($\uparrow$) & $\uparrow$ & In $\uparrow$ & Out 	$\uparrow$  \\ \midrule
 
 Standard SSD & 22.7(50.4) & 84.1 & \textbf{96.8} & 48.4 & 16.2(61.7) & 91.3 & 98.8 & 70.6 & 23.5(50.4) & 85.1 & 98.0 & 52.5 & 21.6(53.0) & 86.5 & 94.0 & 75.5\\ 
 BSAS IoU 0.95 & 22.2(54.2) & 84.8 & 96.7 & 51.0 & 10.5(59.6) & 95.2 & \textbf{99.2} & 83.8 & 19.5(56.6) & 88.5 & \textbf{98.5} & 58.8 & 18.6(56.6) & 89.4 & \textbf{94.8} & 82.0\\
 HDBScan Corner & 21.3(53.7) & 84.8 & 96.5 & 51.5 & 12.7(59.6) & 94.0 & 99.0 & 79.6 & 22.7(56.2) & 85.5 & 98.0 & 54.8 & 19.8(56.2) & 88.0 & 94.0 & 79.4\\
 Hungarian Exponential \& SL & \textbf{20.1(55.1)} & \textbf{86.5} & 96.5 & \textbf{58.2} & 10.8(60.4) & 95.0 & 99.1 & 84.1 & 21.1(60.4) & 7.4 & 98.1 & 59.5 & 18.3(56.7) & 89.7 & 94.2 & 83.7\\
 BSAS IoU 0.95 \& SL &21.6(54.2) & 85.5 & 96.6 & 55.0 & 9.9(59.6) & \textbf{95.4} & \textbf{99.2} & \textbf{86.4} & \textbf{17.8(56.6)} & \textbf{90.0} & 98.4 & \textbf{66.7} & \textbf{17.5(56.6)} & \textbf{90.3} & 94.6 & \textbf{85.1}\\
 BSAS excl. IoU 0.9 \& SL & 20.7(55.9) & 86.2 & 96.6 & 57.9 & \textbf{10.3(61.8)} & 95.2 & 99.1 & 85.7 & 20.2(61.8) & 87.9 & 98.1 & 62.7& 18.2(58.0) & 89.9 & 94.2 & 84.7\\

 \bottomrule
\end{tabular}
}
\end{table*}

\begin{figure}[t]
    \centering
    \includegraphics[width=1\linewidth]{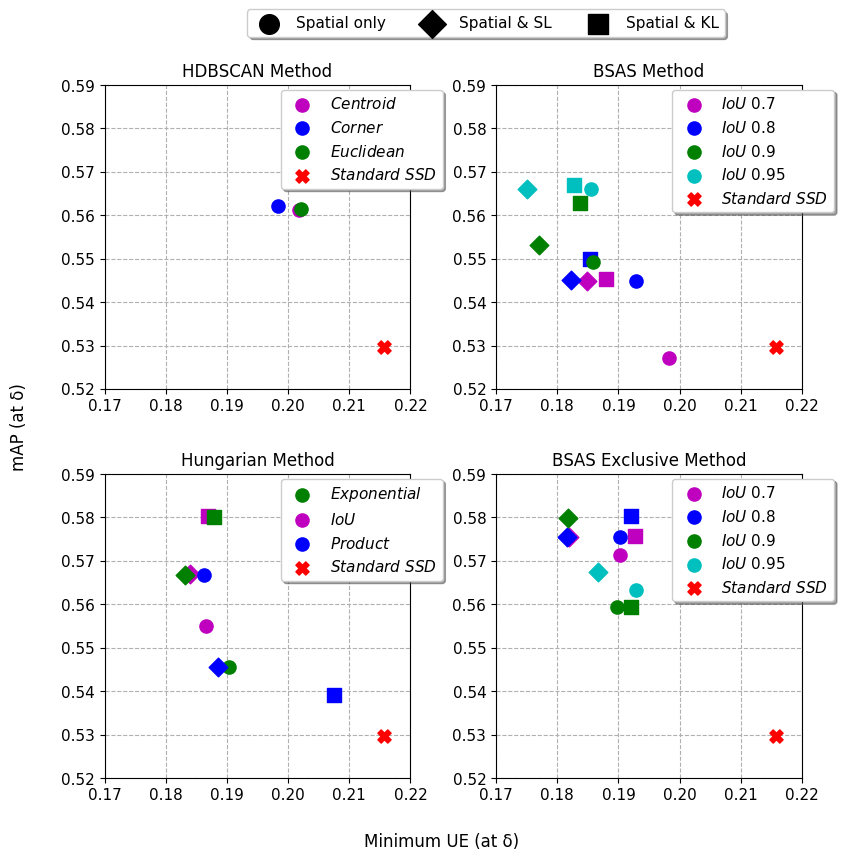}
    \caption{Overall performance of each clustering method and the affinity measures. Best performance is a low minimum uncertainty error (UE) and mAP at least as great as the standard detector.}
    \label{fig:Overall_performance}
\end{figure}
\begin{figure}[tb]
    \centering
    \includegraphics[width=0.9\linewidth]{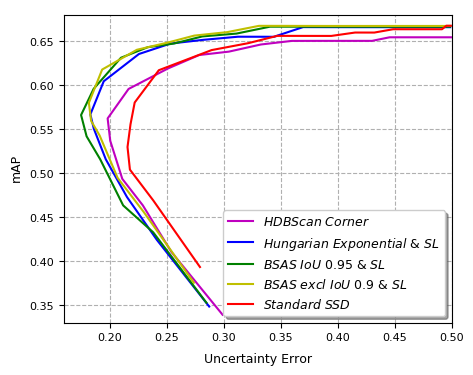}
    \caption{Object detection performance (indicated by mAP) and uncertainty effectiveness (indicated by uncertainty error) for each merging technique. Best performance is in the top left corner, with a high mAP and low detection error.}
    \label{fig:Object_detection_performance}
\end{figure}

\subsection{Summary of Findings}\textbf{Choice of affinity measure and clustering technique substantially affects uncertainty effectiveness and object detection performance}. Between the tested affinity-clustering variations, there is a 3.3\% difference in minimum uncertainty error and a 5.3\% difference in mAP. While all of the clustering methods tested have a higher uncertainty effectiveness than the standard detector, Fig.~\ref{fig:Overall_performance} shows that a poor choice in clustering method can result in lower object detection performance than the standard detector. 

\textbf{BSAS clustering used with affinity measure IoU \& SL and an affinity threshold of 0.95 is the best approach to clustering}. Fig.~\ref{fig:Overall_performance} shows that BSAS IoU 0.95 \& SL is able to achieve the best uncertainty effectiveness (the lowest UE) while also improving object detection performance compared to the standard detector (indicated by mAP). Fig.~\ref{fig:Object_detection_performance} further shows that this method achieves a lower UE than most other clustering methods for the same mAP. 

\textbf{The best affinity measures combine spatial information and a Same Label approach}. As evidenced in Fig.~\ref{fig:Overall_performance}, the majority of clustering methods obtain better performance for uncertainty effectiveness and object detection when combining a spatial affinity with Same Label infinity. This is particularly evident for BSAS, which is not intra-sample exclusive, as distinct objects that have a high spatial overlap can be incorrectly clustered when only spatial affinity is used.  Interestingly, the other semantic affinity, KL, typically offers little improvement and can even result in poorer performance that spatial affinity only. This suggests that the variation introduced by the sampling-based uncertainty technique (MC Dropout) may overwhelm a thorough semantic affinity measure, whereas the general trend between samples (such as the winning label) is more stable.

Section \ref{sec:resultsdepth} offers an extended evaluation on uncertainty effectiveness of the clustering techniques over each of the datasets.

\subsection{Further Review of Performance}
\label{sec:resultsdepth}
Table \ref{tab:performance_detail} shows an in-depth performance evaluation for the best affinity variation of each clustering technique over each of the tested datasets.

\textbf{There is a substantial performance difference between distant open-set and near open-set testing environments. For robotics applications, it is important to evaluate \emph{near open-set} datasets.} When compared to a standard SSD, the worst clustering method (HDBSCAN Corner) demonstrates a 3.5\% decrease in uncertainty error and a 9\% increase in AUPR-Out in distant open-set conditions. However, on the near open-set data, the same method only demonstrates a 0.8\% decrease in uncertainty error and 2.3\% increase in AUPR-Out. While the general trend holds, it is important to recognise that uncertainty evaluated on distant open-set conditions appears to be of higher quality than actual performance in near open-set conditions. This is particularly important for robotics, where a system is likely to operate in near open-set conditions, with an environment that is visually and contextually similar to the training environment but containing objects of unknown classes.

\textbf{AUPR-Out is an important metric for robotic applications where errors can have significant consequences.} For every dataset tested, the difference between the AUPR-In of the standard network and the clustering techniques is negligible. However, there is a notable difference between the AUPR-Out for each clustering strategy. This metric places the most importance on rejecting all incorrect detections, even if some correct detections are also rejected in error. For this reason, it is an important metric to assess for systems where a failure can have serious consequences, such as autonomous systems operating in the real world and interacting with humans and their environment. For any dataset containing open-set data (distant, near or all datasets), BSAS IoU 0.95 \& SL demonstrates the best AUPR-Out score. This is particularly distinct when near open-set conditions are simulated, with BSAS IoU 0.95 \& SL having a $4\%$ increase over the \emph{next best} clustering method (BSAS excl. IoU 0.9 \& SL). 

\textbf{Intra-sample exclusive clustering techniques perform best on closed-set data}. Despite having the highest overall performance, BSAS IoU 0.95 \& SL is outperformed on closed-set data by the methods that enforce intra-sample exclusivity. This is because BSAS IoU 0.95 \& SL is prone to incorrectly clustering detections from the same sample, and while this is not a problem in open-set conditions where all detections are errors, it can decrement performance in closed-set conditions by raising the uncertainty of correct detection. Despite this characteristic of the merging strategy, BSAS IoU 0.95 \& SL still outperforms the standard detector.

\begin{figure}[t]
    \centering
    \includegraphics[width=1\linewidth]{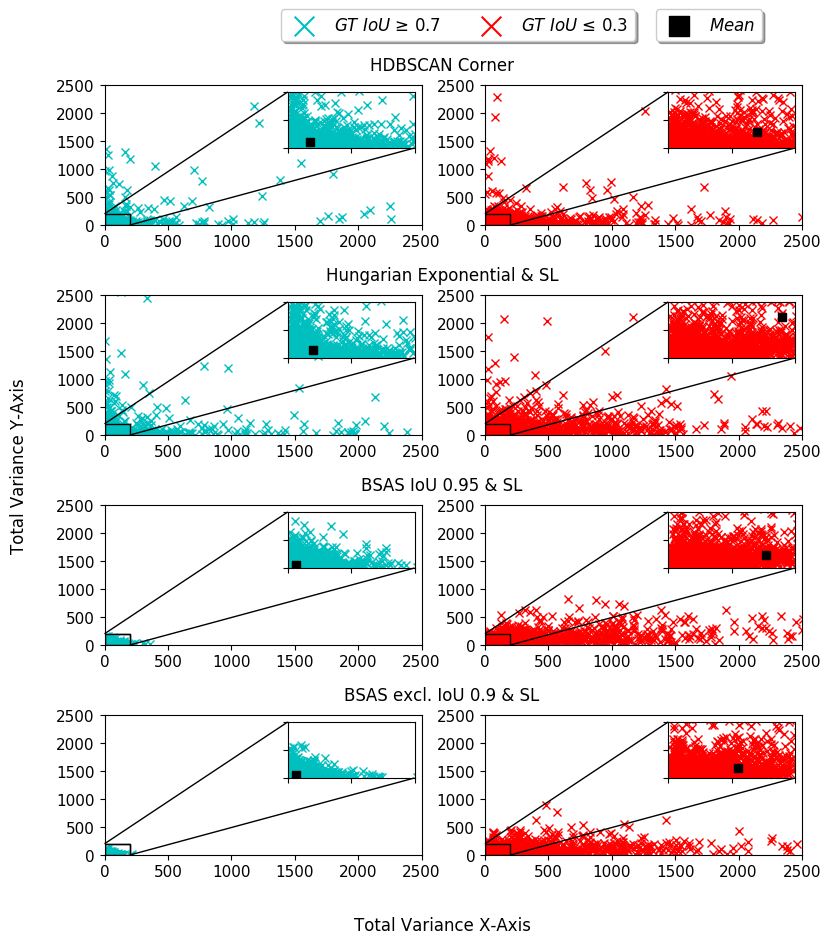}
    \caption{The total variance in pixels (on the x-axis and y-axis) of the detection bounding boxes within an observation cluster. This is compared for observations that have a high spatial accuracy (IoU with a GT object is $\geq$ 0.7) and low spatial accuracy (IoU with a GT object is  $\leq$ 0.3). }
    \label{fig:spatial}
\end{figure}

\textbf{Spatial uncertainty can be extracted from an observation cluster and related to spatial accuracy.} Figure \ref{fig:spatial} shows the total pixel variance in the x-axis and y-axis for the bounding box coordinates within an observation cluster. This total variance can be used to represent the \emph{spatial uncertainty} of an observation. We consider observations with a high spatial accuracy when they have an IoU $\geq 0.7$ with a GT object of the same class and low spatial accuracy when the observations have an IoU $\leq 0.3$ with a GT object of the same class. 

For the BSAS clustering methods, there is a distinct difference in the spread of spatial uncertainties between spatially accurate and inaccurate observations. On the other hand, the HDBSCAN and Hungarian clustering methods have numerous spatially accurate observations that also have a high spatial uncertainty. This is because these methods are threshold-independent, therefore allowing for more errors in clustering where an outlier can introduce greater variance into the cluster. It is important to note that while the BSAS clustering methods do not allow for spatially accurate observations to have high spatial uncertainty, a spatially inaccurate observation can have a low spatial uncertainty. Despite this, there is still an observable difference in the mean spatial uncertainty, thus showing these methods are capable of producing a spatial uncertainty that can be linked to spatial accuracy.

\section{Conclusions and Future Work}\label{sec::conclusions}
In this paper, we investigated the effect of different affinity measures and clustering techniques for correctly associating sample bounding boxes in the context of sampling-based  uncertainty  estimation for object detection. We compared different combinations of three spatial and two semantic affinity measures with four clustering methods. Our results showed that the correct choice of affinity-clustering combinations can greatly improve the effectiveness of the classification and spatial uncertainty estimation and the resulting object detection performance.

\bibliographystyle{IEEEtran}
\bibliography{refs}
\end{document}